\documentclass[runningheads]{llncs}
\usepackage{graphicx}
\usepackage{framed}
\usepackage{placeins}
\usepackage{multirow}

\usepackage[symbol]{footmisc}
\usepackage{amsfonts}

\begin{document}
\title{\textit{GroupEnc}: encoder with group loss for global structure preservation}
\titlerunning{\textit{GroupEnc}: encoder with group loss for global structure preservation}
\author{David Novak\inst{1,2}\orcidID{0000-0003-4574-9093} \and
Sofie Van Gassen\inst{1,2}\orcidID{0000-0002-7119-5330} \and
Yvan Saeys\inst{1,2}\orcidID{0000-0002-0415-1506}}
\authorrunning{Novak et al.}
\institute{Department of Applied Mathematics, Computer Science and Statistics, Ghent University, Belgium \and
Data Mining and Modeling for Biomedicine, Center for Inflammation Research, VIB-UGent, Belgium}
\maketitle
\begin{abstract}
Recent advances in dimensionality reduction have achieved more accurate lower-dimensional embeddings of high-dimensional data.
In addition to visualisation purposes, these embeddings can be used for downstream processing, including batch effect normalisation, clustering, community detection or trajectory inference.
We use the notion of structure preservation at both local and global levels to create a deep learning model, based on a variational autoencoder (VAE) and the stochastic quartet loss from the \textit{SQuadMDS} algorithm.
Our encoder model, called \textit{GroupEnc}, uses a `group loss' function to create embeddings with less global structure distortion than VAEs do, while keeping the model parametric and the architecture flexible.
We validate our approach using publicly available biological single-cell transcriptomic datasets, employing $R_{\mathrm{NX}}$ curves for evaluation.

\end{abstract}
\section{Introduction}

Autoencoders (AEs) are neural networks which encode high-dimensional (HD) input data as a low-dimensional (LD) latent representation and decode this into a reconstruction of the input.
In training, reconstruction error is minimised via back-propagation.

In the field of bioinformatics, we have seen impressive applications of autoencoders and variational autoencoders (VAEs; probabilistic models based on AEs) in dimensionality reduction (DR) for the purposes of visualisation \cite{Szubert2019,Ding2018} and downstream data processing, including batch effect correction and cell population clustering \cite{Amodio2019,Chen2020,Kopf2021}.
This pertains to large and high-dimensional single-cell datasets, which quantify biological features per cell in a tissue sample of interest.
Examples of these methods include single-cell RNA sequencing (scRNA-seq), flow cytometry, mass cytometry (\textit{CyTOF}) and CITE-seq.

We introduce and evaluate \textit{GroupEnc}: a stand-alone encoder module that optimises the \textit{group loss}: a differentiable loss function that imposes a scale-agnostic structure-preserving constraint on the learned LD embedding.
This is a modification of the stochastic quartet loss in \textit{SQuadMDS} \cite{Lambert2021}, applied in a deep learning context here.
This results in a parametric model that can run on GPU.
We achieve similar local structure preservation and better global structure preservation than a VAE model, as tested on 5 single-cell transcriptomic datasets.
Compared to previously published alternative triplet-based loss functions proposed for VAEs \cite{Szubert2019,Amid2022}, the group loss does not require computation of a \textit{k}-nearest-neighbour graph of the input data.

\section{Method}

We describe the methodology used to create LD embeddings of HD data and to evaluate them.

\subsection{Model training}

In an autoencoder architecture, HD input $\mathbf{X_{n\times d}} \in \mathcal{X}$ is encoded as LD representation $\mathbf{Z_{n\times w}} \in \mathcal{Z}$ (where $\mathcal{X}=\mathbb{R}^{d}, \mathcal{Z}=\mathbb{R}^{w}, w<d$) and reconstructed as an approximation $\mathbf{\hat{X}_{n\times d}}$.

The encoder $E_{\Phi}: \mathcal{X} \rightarrow \mathcal{Z}$ transforms $\mathbf{X}$ to $\mathbf{L}$, and the decoder $D_{\theta}: \mathcal{Z} \rightarrow \mathcal{X}$ transforms $\mathbf{L}$ to $\mathbf{\hat{X}}$.
Parameters of the AE (encoder weights $\Phi$ and decoder weights $\Theta$) are learned so as to reduce a reconstruction loss.
In our baseline VAE model, we use the mean square error (MSE) as reconstruction loss.

In a VAE, the latent representation $\mathbf{L}$ is sampled from a distribution $\mathcal{D}$ in latent space.
The encoder and decoder networks are probabilistic, and an extra term quantifying the Kullback-Leibler (KL) divergence between $\mathcal{D}$ and a latent prior (isotropic Gaussian distribution) is used as an additional loss term during training.

In contrast, our current \textit{GroupEnc} model only consists of a variational encoder and sampler (without a decoder), trained to minimise a \textit{group loss} along with the KL divergence from prior.
The group loss adapts the notion of the quartet loss function, computed using quartet-normalised distances between original and embedded points, from \textit{SQuadMDS} \cite{Lambert2021}.
The normalised distances are used to calculate a differentiable cost function per each randomly drawn quartet of points.
We denote Euclidean distances between any HD input points or LD embedded points indexed $i$ and $j$ as $\delta_{ij}$ and $d_{ij}$, respectively.
To compute a group-normalised distance between two points in the same group (for a quartet, quintet, sextet, etc.), we use all pairwise distances within that group.
For HD and LD points, respectively, we get group-normalised distance formulas \begin{equation}
\delta^{\mathrm{norm}}_{ij} = \frac{\delta_{ij}}{\sum^{\gamma-1}_{a=1}\sum^{\gamma}_{b=a+1}\delta_{ab}}\end{equation} \begin{equation}d^{\mathrm{norm}}_{ij} = \frac{d_{ij}}{\sum^{\gamma-1}_{a=1}\sum^{\gamma}_{b=a+1}d_{ab}}\end{equation} where $\gamma$ is the number of points in each group.

\begin{figure}
  \begin{framed}
    \centering
    \includegraphics[width=320px]{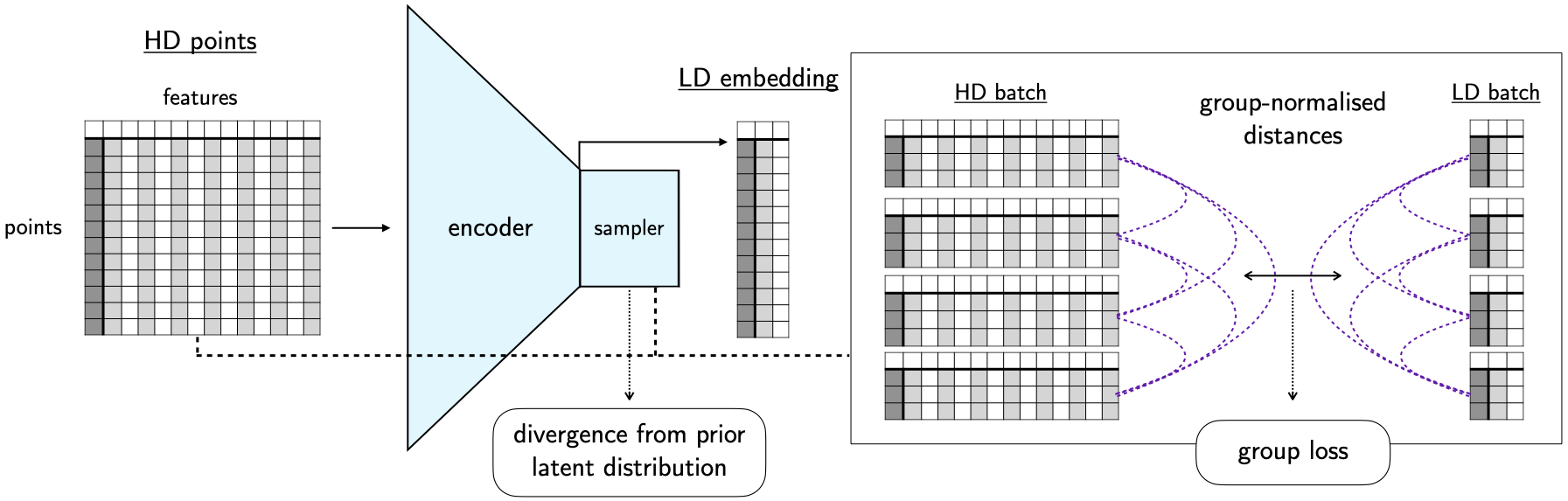}
    \caption[fig1]{A schematic illustration of \textit{GroupEnc} training and inference is shown. In training, a batch of high-dimensional points is used as input to the encoder. Parameters of the model are adjusted in each pass via back-propagation to minimise the group loss, which quantifies divergence in relative distances within randomly assigned groups of points in a batch of input points versus its embedding. The trained encoder then outputs a low-dimensional embeddding of the input.}
    \label{fig1}
  \end{framed}
\end{figure}

The difference in group-normalised distances in HD and LD, which ought to be minimised, is used to calculate the cost function \begin{equation}g = \sum\limits^{\gamma-1}_{a=1}\sum\limits^{\gamma}_{b=a+1}(\delta^{\mathrm{norm}}_{ab}-d^{\mathrm{norm}}_{ab})^{2}\end{equation} of a group (a \textit{group cost}). This is visualised in Figure \ref{fig1}.

The \textit{GroupEnc} model is trained on shuffled batches of input data using the \textit{Adam} optimiser.
Partitioning of points into groups is done dynamically at the batch level, and the size of the groups ($\gamma$) is specified as a hyperparameter.
The group loss value per each point \textit{i} in the training batch is assigned as the cost value of the group for which \textit{i} is the first point, and the group loss term per batch is averaged across the batch.

Therefore, \textit{GroupEnc} imposes a constraint on the latent distribution $\mathcal{D}$ instead of reconstruction loss to compute weight updates.

\subsection{Dimensionality reduction quality assessment}

To assess structure preservation (SP) in an embedding, we use the $R_{\mathrm{NX}}$ curve, a previously proposed quality assessment metric \cite{Lee2015}.
This curve quantifies the overlap between ordering of neighbours to a reference point in HD versus in LD for all neighbourhood sizes, from $1$ to $(N-1)$ (with sample size $N$), averaged across all reference points.

To compute this, we denote neighbourhood ranks of a point $j$ (neighbour) with respect to a point $i$ (reference point) as $\rho_{ij}$ and $r_{ij}$ in HD and in LD, respectively.
Non-self neighbourhoods of HD and LD points, respectively, are then denoted as $\nu_{i}^{K}=\{ j: 1 \leq \rho_{ij} \leq K \}$ and $n_{i}^{K}=\{ j: 1 \leq r_{ij} \leq K \}$ for neighbourhood size $K$.
For dataset size $N$, the $Q_{\mathrm{NX}}$ value for a specific value of $K$ is calculated as \begin{equation}Q_{\mathrm{NX}}(K)=\frac{1}{KN}\sum\limits_{i=1}^{N}\vert \nu_{i}^{K} \cap n_{i}^{K} \vert\end{equation}
To obtain the full $Q_{\mathrm{NX}}$ curve, we calculate this score for $K$ from 1 to $(N-1)$.

It turns out that a random embedding results in $Q_{\mathrm{NX}}(K) \approx \frac{K}{N-1}$. 
$R_{\mathrm{NX}}$, as opposed to $Q_{\mathrm{NX}}$, corrects for chance, and is computed as \begin{equation}R_{\mathrm{NX}}(K)=\frac{(N-1)Q_{\mathrm{NX}}(K)-K}{N-1-K}\end{equation}

We quantify SP as the area-under-curve (AUC) for an $R_{\mathrm{NX}}$ curve of an embedding of interest.
Specifically, Local SP is the AUC of the curve where neighbourhood size ($K$) is re-scaled logarithmically ($ln K$ is used), to up-weight local neighbourhoods while not setting a hard cut-off for local versus global.
Moreover, Global SP is the AUC with a linear scale for $K$, therefore without the emphasis on local neighbourhoods of the reference points.
In both cases, a higher SP score is better.

\section{Results}

We compare a VAE (trained to minimise reconstruction error and KL-divergence from prior) and a \textit{GroupEnc} model (encoder-only, trained to minimise group loss)\footnote[1]{The encoder module, in both cases, consisted of layers sized $(32, 64, 128, 32)$ and the VAE decoder module of layers consisted of layers sized $(32, 128, 64, 32)$.
The \textit{Adam} optimiser with a learning rate of $0.001$ was used for $500$ epochs of training with batch size of $512$.}.

We tested structure preservation (SP) in embeddings of dimensionality 2, 5 and 10, with different values of hyperparameter $\gamma$ (group size), looking at Local and Global SP separately.

We use 5 single-cell RNA-sequencing (scRNA-seq) datasets \cite{Liu2023,Farrell2019,Shekhar2016,Ximerakis2019,Ziegler2021}, comprising high-dimensional feature vectors describing the identity of single biological cells in a tissue sample of interest.
These features are levels of transcription of labelled genes.
The datasets are listed in Table \ref{table_datasets}.

\begin{figure}
  \begin{framed}
    \centering
    \includegraphics[width=300px]{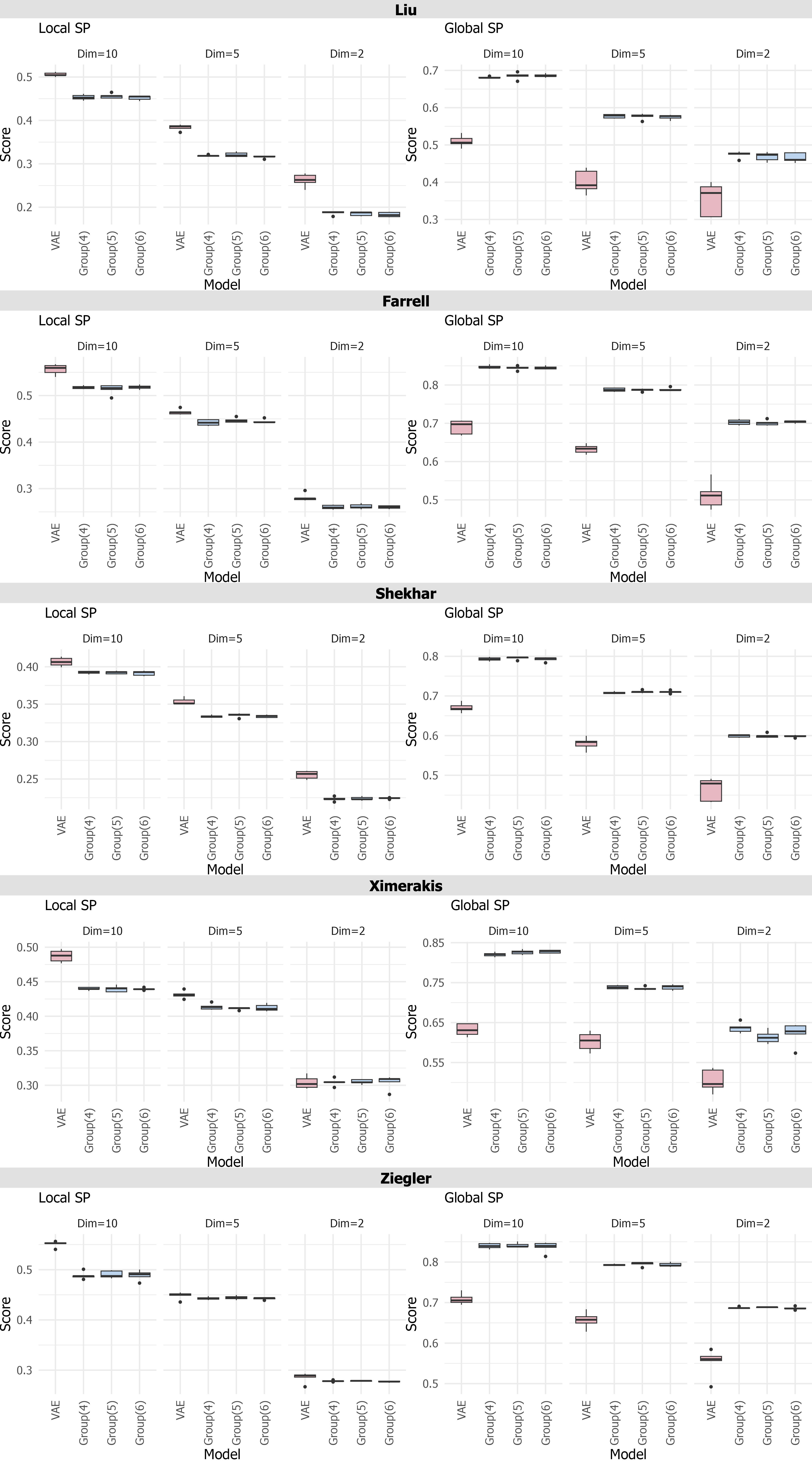}
    \caption[fig_boxplots]{Boxplots of Local and Global SP scores across 5 runs for embeddings of each dataset, obtained from VAE and \textit{GroupEnc} model for different group size ($\gamma$) values (written as `Group($\gamma$)'). Subplots are sorted by target dimensionalities (columns).}
    \label{fig_boxplots}
  \end{framed}
\end{figure}

Local SP and Global SP scores are summarised in Figure \ref{fig_boxplots} and shown in Tables 2 and 3 in full. Time required to train each model can be found in Table 4, with a single node of a GPU cluster (\textit{16-core Intel Xeon Gold 6242} processor with \textit{NVIDIA Volta V100} GPU) with 16 GB of usable RAM made available each time.
5 runs (with different random seeds) were run to collect the scores.

\FloatBarrier

\FloatBarrier
{\renewcommand{\arraystretch}{1.0}
\begin{table}[!htbp]
\centering
\label{table_datasets}
\fontsize{8}{11}\selectfont
\setlength{\tabcolsep}{0.5em}
\begin{tabular}{|l|l|l|l|l|}
\hline
\scalebox{.8}[1.0]{\textbf{Dataset name}} & \scalebox{.8}[1.0]{\textbf{Biological source}} & \scalebox{.8}[1.0]{\textbf{Feature set}} & \scalebox{.8}[1.0]{\textbf{Number of samples}} \\
\hline
Ziegler & human nasopharynx & 32,871 genes & 32,588 cells \\
\hline
Shekhar & mouse retina & 24,904 genes & 44,994 cells \\
\hline
Ximerakis & mouse brain & 14,699 genes & 37,069 cells \\
\hline
Farrell & zebrafish embryos & 17,239 genes & 38,731 cells \\
\hline
Liu & mouse brain & 17,482 genes & 26,187 cells \\
\hline
\end{tabular}
\begin{center}
\caption{Datasets used for DR benchmark and their brief descriptions.}
\label{table_datasets}
\end{center}
\centering
\fontsize{8}{11}\selectfont
\setlength{\tabcolsep}{0.5em}
\begin{tabular}{|l|l|lllll|}
\hline
\multirow{2}{*}{\scalebox{.8}[1.0]{\textbf{Dim}}} & \multirow{2}{*}{\scalebox{.8}[1.0]{\textbf{Model}}} & \multicolumn{5}{l|}{\scalebox{.8}[1.0]{\textbf{Dataset}}}                                                                                                                                                                                                                      \\ \cline{3-7} 
                                &                        & \multicolumn{1}{l|}{Liu}                      & \multicolumn{1}{l|}{Farrell}                       & \multicolumn{1}{l|}{Shekhar}                       & \multicolumn{1}{l|}{Ximerakis}                     & Ziegler                       \\ \hline
\multirow{4}{*}{$10d$}            & VAE                    & \multicolumn{1}{l|}{$0.506{\scriptstyle\pm0.005}$} & \multicolumn{1}{l|}{$0.556{\scriptstyle\pm0.011}$} & \multicolumn{1}{l|}{$0.407{\scriptstyle\pm0.006}$} & \multicolumn{1}{l|}{$0.487{\scriptstyle\pm0.009}$} & $0.551{\scriptstyle\pm0.006}$ \\ \cline{2-7} 
                                & GroupEnc ($\gamma=4$)          & \multicolumn{1}{l|}{$0.453{\scriptstyle\pm0.006}$} & \multicolumn{1}{l|}{$0.518{\scriptstyle\pm0.003}$} & \multicolumn{1}{l|}{$0.392{\scriptstyle\pm0.002}$} & \multicolumn{1}{l|}{$0.440{\scriptstyle\pm0.002}$} & $0.488{\scriptstyle\pm0.007}$ \\ \cline{2-7} 
                                & GroupEnc ($\gamma=5$)         & \multicolumn{1}{l|}{$0.456{\scriptstyle\pm0.006}$} & \multicolumn{1}{l|}{$0.513{\scriptstyle\pm0.011}$} & \multicolumn{1}{l|}{$0.392{\scriptstyle\pm0.002}$} & \multicolumn{1}{l|}{$0.439{\scriptstyle\pm0.005}$} & $0.490{\scriptstyle\pm0.007}$ \\ \cline{2-7} 
                                & GroupEnc ($\gamma=6$)         & \multicolumn{1}{l|}{$0.452{\scriptstyle\pm0.005}$} & \multicolumn{1}{l|}{$0.518{\scriptstyle\pm0.004}$} & \multicolumn{1}{l|}{$0.391{\scriptstyle\pm0.003}$} & \multicolumn{1}{l|}{$0.439{\scriptstyle\pm0.002}$} & $0.489{\scriptstyle\pm0.010}$ \\ \hline \hline
\multirow{4}{*}{$5d$}             & VAE                    & \multicolumn{1}{l|}{$0.384{\scriptstyle\pm0.007}$} & \multicolumn{1}{l|}{$0.465{\scriptstyle\pm0.006}$} & \multicolumn{1}{l|}{$0.354{\scriptstyle\pm0.004}$} & \multicolumn{1}{l|}{$0.431{\scriptstyle\pm0.005}$} & $0.448{\scriptstyle\pm0.007}$ \\ \cline{2-7} 
                                & GroupEnc ($\gamma=4$)          & \multicolumn{1}{l|}{$0.318{\scriptstyle\pm0.002}$} & \multicolumn{1}{l|}{$0.442{\scriptstyle\pm0.007}$} & \multicolumn{1}{l|}{$0.334{\scriptstyle\pm0.002}$} & \multicolumn{1}{l|}{$0.414{\scriptstyle\pm0.004}$} & $0.443{\scriptstyle\pm0.003}$ \\ \cline{2-7} 
                                & GroupEnc ($\gamma=5$)          & \multicolumn{1}{l|}{$0.321{\scriptstyle\pm0.005}$} & \multicolumn{1}{l|}{$0.447{\scriptstyle\pm0.005}$} & \multicolumn{1}{l|}{$0.335{\scriptstyle\pm0.003}$} & \multicolumn{1}{l|}{$0.411{\scriptstyle\pm0.002}$} & $0.444{\scriptstyle\pm0.004}$ \\ \cline{2-7} 
                                & GroupEnc ($\gamma=6$)          & \multicolumn{1}{l|}{$0.316{\scriptstyle\pm0.003}$} & \multicolumn{1}{l|}{$0.444{\scriptstyle\pm0.004}$} & \multicolumn{1}{l|}{$0.334{\scriptstyle\pm0.002}$} & \multicolumn{1}{l|}{$0.412{\scriptstyle\pm0.005}$} & $0.443{\scriptstyle\pm0.002}$ \\ \hline \hline
\multirow{4}{*}{$2d$}             & VAE                    & \multicolumn{1}{l|}{$0.262{\scriptstyle\pm0.015}$} & \multicolumn{1}{l|}{$0.281{\scriptstyle\pm0.009}$} & \multicolumn{1}{l|}{$0.256{\scriptstyle\pm0.005}$} & \multicolumn{1}{l|}{$0.304{\scriptstyle\pm0.009}$} & $0.285{\scriptstyle\pm0.010}$ \\ \cline{2-7} 
                                & GroupEnc ($\gamma=4$)          & \multicolumn{1}{l|}{$0.187{\scriptstyle\pm0.005}$} & \multicolumn{1}{l|}{$0.260{\scriptstyle\pm0.005}$} & \multicolumn{1}{l|}{$0.223{\scriptstyle\pm0.003}$} & \multicolumn{1}{l|}{$0.304{\scriptstyle\pm0.005}$} & $0.278{\scriptstyle\pm0.001}$ \\ \cline{2-7} 
                                & GroupEnc ($\gamma=5$)          & \multicolumn{1}{l|}{$0.185{\scriptstyle\pm0.005}$} & \multicolumn{1}{l|}{$0.262{\scriptstyle\pm0.005}$} & \multicolumn{1}{l|}{$0.224{\scriptstyle\pm0.002}$} & \multicolumn{1}{l|}{$0.305{\scriptstyle\pm0.003}$} & $0.278{\scriptstyle\pm0.001}$ \\ \cline{2-7} 
                                & GroupEnc ($\gamma=6$)          & \multicolumn{1}{l|}{$0.183{\scriptstyle\pm0.006}$} & \multicolumn{1}{l|}{$0.260{\scriptstyle\pm0.003}$} & \multicolumn{1}{l|}{$0.224{\scriptstyle\pm0.001}$} & \multicolumn{1}{l|}{$0.304{\scriptstyle\pm0.010}$} & $0.277{\scriptstyle\pm0.001}$ \\ \hline
\end{tabular}
\begin{center}
\caption{Local SP for 5 datasets, 3 embedding dimensionalities (`Dim') and 4 models (VAE and \textit{GroupEnc} with group size $\gamma$ of 4, 5 and 6). Mean and standard deviation are shown.}
\end{center}
\label{table_scores_local}
\centering
\fontsize{8}{11}\selectfont
\setlength{\tabcolsep}{0.5em}
\begin{tabular}{|r|l|lllll|}
\hline
\multirow{2}{*}{\scalebox{.8}[1.0]{\textbf{Dim}}} & \multirow{2}{*}{\scalebox{.8}[1.0]{\textbf{Model}}} & \multicolumn{5}{l|}{\scalebox{.8}[1.0]{\textbf{Dataset}}}                                                                                                                                                                                                                      \\ \cline{3-7} 
\multicolumn{1}{|l|}{}                                &                        & \multicolumn{1}{l|}{Liu}                      & \multicolumn{1}{l|}{Farrell}                       & \multicolumn{1}{l|}{Shekhar}                       & \multicolumn{1}{l|}{Ximerakis}                     & Ziegler                       \\ \hline
\multirow{4}{*}{$10d$}                                  & VAE                    & \multicolumn{1}{l|}{$0.510{\scriptstyle\pm0.016}$} & \multicolumn{1}{l|}{$0.690{\scriptstyle\pm0.019}$} & \multicolumn{1}{l|}{$0.670{\scriptstyle\pm0.012}$} & \multicolumn{1}{l|}{$0.632{\scriptstyle\pm0.015}$} & $0.709{\scriptstyle\pm0.014}$ \\ \cline{2-7} 
                                                      & GroupEnc ($\gamma=4$)          & \multicolumn{1}{l|}{$0.681{\scriptstyle\pm0.002}$} & \multicolumn{1}{l|}{$0.847{\scriptstyle\pm0.005}$} & \multicolumn{1}{l|}{$0.793{\scriptstyle\pm0.005}$} & \multicolumn{1}{l|}{$0.820{\scriptstyle\pm0.005}$} & $0.840{\scriptstyle\pm0.007}$ \\ \cline{2-7} 
                                                      & GroupEnc ($\gamma=5$)         & \multicolumn{1}{l|}{$0.685{\scriptstyle\pm0.009}$} & \multicolumn{1}{l|}{$0.844{\scriptstyle\pm0.005}$} & \multicolumn{1}{l|}{$0.796{\scriptstyle\pm0.004}$} & \multicolumn{1}{l|}{$0.826{\scriptstyle\pm0.006}$} & $0.841{\scriptstyle\pm0.006}$ \\ \cline{2-7} 
                                                      & GroupEnc ($\gamma=6$)         & \multicolumn{1}{l|}{$0.686{\scriptstyle\pm0.005}$} & \multicolumn{1}{l|}{$0.845{\scriptstyle\pm0.004}$} & \multicolumn{1}{l|}{$0.793{\scriptstyle\pm0.005}$} & \multicolumn{1}{l|}{$0.828{\scriptstyle\pm0.004}$} & $0.837{\scriptstyle\pm0.014}$ \\ \hline \hline
\multirow{4}{*}{$5d$}                                   & VAE                    & \multicolumn{1}{l|}{$0.401{\scriptstyle\pm0.032}$} & \multicolumn{1}{l|}{$0.633{\scriptstyle\pm0.012}$} & \multicolumn{1}{l|}{$0.580{\scriptstyle\pm0.016}$} & \multicolumn{1}{l|}{$0.602{\scriptstyle\pm0.024}$} & $0.657{\scriptstyle\pm0.020}$ \\ \cline{2-7} 
                                                      & GroupEnc ($\gamma=4$)          & \multicolumn{1}{l|}{$0.577{\scriptstyle\pm0.005}$} & \multicolumn{1}{l|}{$0.788{\scriptstyle\pm0.005}$} & \multicolumn{1}{l|}{$0.708{\scriptstyle\pm0.003}$} & \multicolumn{1}{l|}{$0.739{\scriptstyle\pm0.005}$} & $0.793{\scriptstyle\pm0.002}$ \\ \cline{2-7} 
                                                      & GroupEnc ($\gamma=5$)          & \multicolumn{1}{l|}{$0.576{\scriptstyle\pm0.008}$} & \multicolumn{1}{l|}{$0.787{\scriptstyle\pm0.003}$} & \multicolumn{1}{l|}{$0.710{\scriptstyle\pm0.003}$} & \multicolumn{1}{l|}{$0.735{\scriptstyle\pm0.004}$} & $0.795{\scriptstyle\pm0.005}$ \\ \cline{2-7} 
                                                      & GroupEnc ($\gamma=6$)          & \multicolumn{1}{l|}{$0.575{\scriptstyle\pm0.007}$} & \multicolumn{1}{l|}{$0.788{\scriptstyle\pm0.004}$} & \multicolumn{1}{l|}{$0.710{\scriptstyle\pm0.003}$} & \multicolumn{1}{l|}{$0.738{\scriptstyle\pm0.007}$} & $0.793{\scriptstyle\pm0.005}$ \\ \hline \hline
\multirow{4}{*}{$2d$}                                   & VAE                    & \multicolumn{1}{l|}{$0.355{\scriptstyle\pm0.045}$} & \multicolumn{1}{l|}{$0.512{\scriptstyle\pm0.036}$} & \multicolumn{1}{l|}{$0.465{\scriptstyle\pm0.029}$} & \multicolumn{1}{l|}{$0.504{\scriptstyle\pm0.028}$} & $0.552{\scriptstyle\pm0.035}$ \\ \cline{2-7} 
                                                      & GroupEnc ($\gamma=4$)          & \multicolumn{1}{l|}{$0.474{\scriptstyle\pm0.009}$} & \multicolumn{1}{l|}{$0.703{\scriptstyle\pm0.007}$} & \multicolumn{1}{l|}{$0.599{\scriptstyle\pm0.003}$} & \multicolumn{1}{l|}{$0.637{\scriptstyle\pm0.013}$} & $0.687{\scriptstyle\pm0.002}$ \\ \cline{2-7} 
                                                      & GroupEnc ($\gamma=5$)          & \multicolumn{1}{l|}{$0.469{\scriptstyle\pm0.012}$} & \multicolumn{1}{l|}{$0.701{\scriptstyle\pm0.007}$} & \multicolumn{1}{l|}{$0.599{\scriptstyle\pm0.005}$} & \multicolumn{1}{l|}{$0.614{\scriptstyle\pm0.016}$} & $0.689{\scriptstyle\pm0.001}$ \\ \cline{2-7} 
                                                      & GroupEnc ($\gamma=6$)          & \multicolumn{1}{l|}{$0.466{\scriptstyle\pm0.013}$} & \multicolumn{1}{l|}{$0.704{\scriptstyle\pm0.003}$} & \multicolumn{1}{l|}{$0.598{\scriptstyle\pm0.003}$} & \multicolumn{1}{l|}{$0.622{\scriptstyle\pm0.028}$} & $0.686{\scriptstyle\pm0.004}$ \\ \hline
\end{tabular}
\begin{center}
\caption{Global SP for 5 datasets, 3 embedding dimensionalities (`Dim') and 4 models (VAE and \textit{GroupEnc} with group size $\gamma$ of 4, 5 and 6). Mean and standard deviation are shown.}
\end{center}
\label{table_scores_global}
\end{table}
}
\FloatBarrier

\begin{table}[]
\centering
\fontsize{8}{11}\selectfont
\setlength{\tabcolsep}{0.5em}
\begin{tabular}{|r|l|lllll|}
\hline
\multirow{2}{*}{\scalebox{.8}[1.0]{\textbf{Dim}}} & \multirow{2}{*}{\scalebox{.8}[1.0]{\textbf{Model}}} & \multicolumn{5}{l|}{\scalebox{.8}[1.0]{\textbf{Dataset}}}                                                                                                                                                                                                                 \\ \cline{3-7} 
\multicolumn{1}{|l|}{}                                &                        & \multicolumn{1}{l|}{Campbell}                     & \multicolumn{1}{l|}{Farrell}                      & \multicolumn{1}{l|}{Shekhar}                      & \multicolumn{1}{l|}{Ximerakis}                    & Ziegler                      \\ \hline
\multirow{4}{*}{$10d$}                                  & VAE                    & \multicolumn{1}{l|}{$89.7{\scriptstyle\pm9.4}$}   & \multicolumn{1}{l|}{$126.2{\scriptstyle\pm15.9}$} & \multicolumn{1}{l|}{$138.2{\scriptstyle\pm15.8}$} & \multicolumn{1}{l|}{$116.4{\scriptstyle\pm11.2}$} & $103.9{\scriptstyle\pm11.3}$ \\ \cline{2-7} 
                                                      & GroupEnc ($\gamma=4$)          & \multicolumn{1}{l|}{$151.6{\scriptstyle\pm18.2}$} & \multicolumn{1}{l|}{$221.8{\scriptstyle\pm19.1}$} & \multicolumn{1}{l|}{$263.0{\scriptstyle\pm35.2}$} & \multicolumn{1}{l|}{$211.6{\scriptstyle\pm20.3}$} & $187.9{\scriptstyle\pm24.0}$ \\ \cline{2-7} 
                                                      & GroupEnc ($\gamma=5$)         & \multicolumn{1}{l|}{$150.2{\scriptstyle\pm15.5}$} & \multicolumn{1}{l|}{$228.7{\scriptstyle\pm27.0}$} & \multicolumn{1}{l|}{$252.1{\scriptstyle\pm25.6}$} & \multicolumn{1}{l|}{$222.9{\scriptstyle\pm23.2}$} & $172.9{\scriptstyle\pm5.3}$  \\ \cline{2-7} 
                                                       & GroupEnc ($\gamma=6$)         & \multicolumn{1}{l|}{$156.7{\scriptstyle\pm20.0}$} & \multicolumn{1}{l|}{$221.9{\scriptstyle\pm23.8}$} & \multicolumn{1}{l|}{$252.6{\scriptstyle\pm28.6}$} & \multicolumn{1}{l|}{$213.1{\scriptstyle\pm21.4}$} & $189.5{\scriptstyle\pm22.1}$ \\ \hline \hline
\multirow{4}{*}{$5d$}                                   & VAE                    & \multicolumn{1}{l|}{$78.9{\scriptstyle\pm3.5}$}   & \multicolumn{1}{l|}{$121.9{\scriptstyle\pm10.8}$} & \multicolumn{1}{l|}{$137.8{\scriptstyle\pm15.4}$} & \multicolumn{1}{l|}{$120.1{\scriptstyle\pm12.9}$} & $100.2{\scriptstyle\pm6.7}$  \\ \cline{2-7} 
                                                      & GroupEnc ($\gamma=4$)          & \multicolumn{1}{l|}{$155.9{\scriptstyle\pm20.0}$} & \multicolumn{1}{l|}{$227.2{\scriptstyle\pm29.2}$} & \multicolumn{1}{l|}{$258.6{\scriptstyle\pm34.7}$} & \multicolumn{1}{l|}{$207.2{\scriptstyle\pm24.0}$} & $179.9{\scriptstyle\pm17.4}$ \\ \cline{2-7} 
                                                      & GroupEnc ($\gamma=5$)          & \multicolumn{1}{l|}{$148.5{\scriptstyle\pm17.1}$} & \multicolumn{1}{l|}{$226.7{\scriptstyle\pm28.9}$} & \multicolumn{1}{l|}{$249.9{\scriptstyle\pm30.2}$} & \multicolumn{1}{l|}{$207.8{\scriptstyle\pm24.4}$} & $177.3{\scriptstyle\pm18.3}$ \\ \cline{2-7} 
                                                      & GroupEnc ($\gamma=6$)          & \multicolumn{1}{l|}{$150.0{\scriptstyle\pm21.0}$} & \multicolumn{1}{l|}{$227.2{\scriptstyle\pm28.2}$} & \multicolumn{1}{l|}{$251.3{\scriptstyle\pm26.4}$} & \multicolumn{1}{l|}{$218.1{\scriptstyle\pm27.8}$} & $178.1{\scriptstyle\pm18.1}$ \\ \hline \hline 
\multirow{4}{*}{$2d$}                                   & VAE                    & \multicolumn{1}{l|}{$85.4{\scriptstyle\pm1.6}$}   & \multicolumn{1}{l|}{$123.3{\scriptstyle\pm10.7}$} & \multicolumn{1}{l|}{$149.5{\scriptstyle\pm12.9}$} & \multicolumn{1}{l|}{$117.1{\scriptstyle\pm12.2}$} & $100.1{\scriptstyle\pm11.1}$ \\ \cline{2-7} 
                                                      & GroupEnc ($\gamma=4$)          & \multicolumn{1}{l|}{$151.2{\scriptstyle\pm2.2}$}  & \multicolumn{1}{l|}{$226.8{\scriptstyle\pm27.3}$} & \multicolumn{1}{l|}{$282.0{\scriptstyle\pm27.1}$} & \multicolumn{1}{l|}{$209.0{\scriptstyle\pm21.3}$} & $181.6{\scriptstyle\pm18.0}$ \\ \cline{2-7} 
                                                      & GroupEnc ($\gamma=5$)          & \multicolumn{1}{l|}{$159.7{\scriptstyle\pm16.0}$} & \multicolumn{1}{l|}{$217.3{\scriptstyle\pm24.0}$} & \multicolumn{1}{l|}{$255.6{\scriptstyle\pm23.5}$} & \multicolumn{1}{l|}{$209.0{\scriptstyle\pm22.7}$} & $179.1{\scriptstyle\pm17.3}$ \\ \cline{2-7} 
                                                      & GroupEnc ($\gamma=6$)          & \multicolumn{1}{l|}{$154.4{\scriptstyle\pm13.7}$} & \multicolumn{1}{l|}{$218.0{\scriptstyle\pm24.9}$} & \multicolumn{1}{l|}{$280.1{\scriptstyle\pm31.2}$} & \multicolumn{1}{l|}{$218.0{\scriptstyle\pm30.0}$} & $180.7{\scriptstyle\pm16.4}$ \\ \hline
\end{tabular}
\begin{center}
\caption{Model training time in seconds across 5 datasets, 3 embedding dimensionalities (`Dim') and 4 models (VAE and \textit{GroupEnc} with group size $\gamma$ of 4, 5 and 6). Mean and standard deviation are shown.}
\end{center}
\label{table_times}
\vspace{-1.2cm}
\end{table}

\begin{figure}[!htbp]
  \begin{framed}
    \centering
    \includegraphics[width=280px]{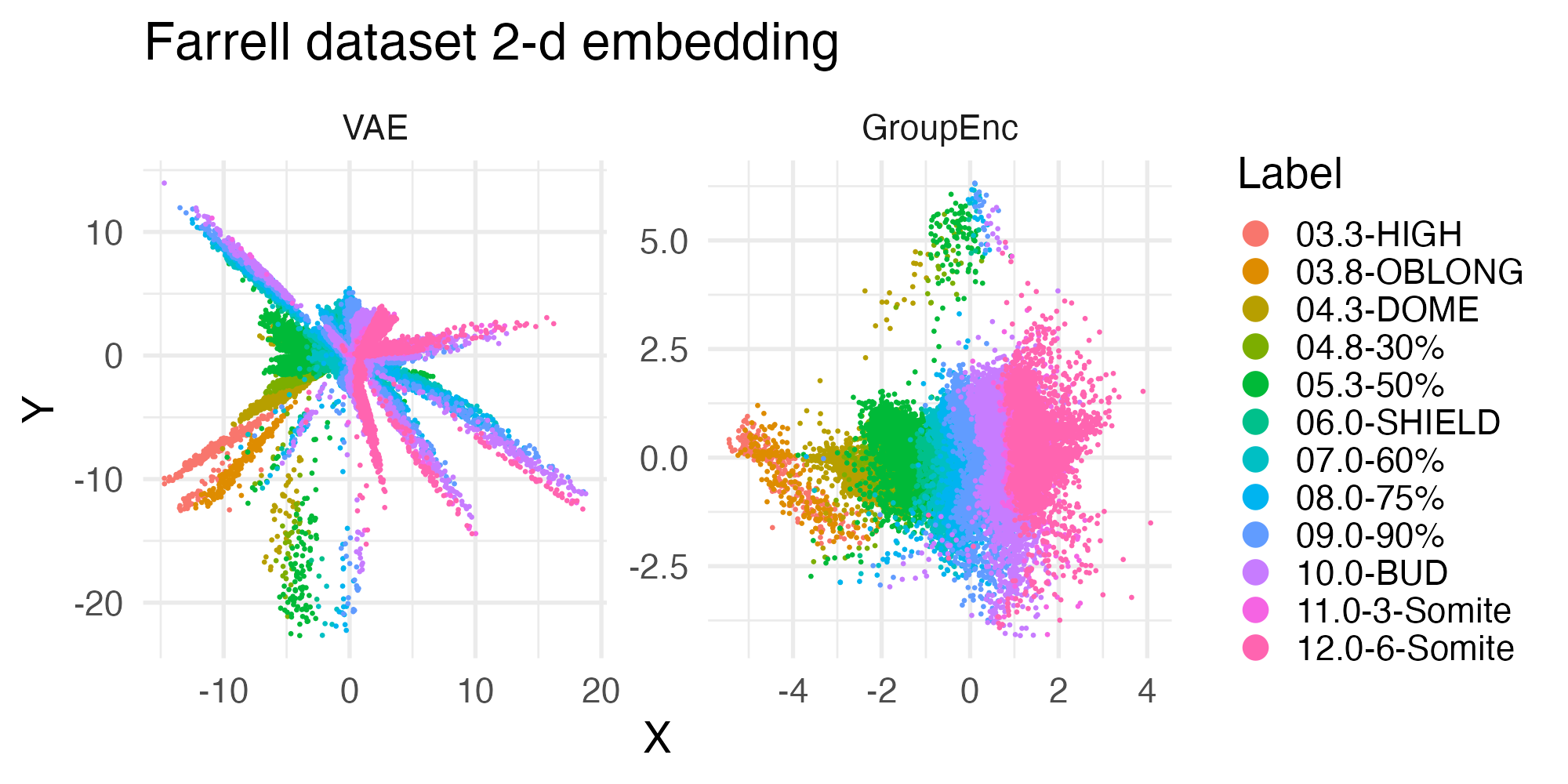}
    \caption[fig_2d]{2-dimensional embeddings of the Farrell dataset obtained using VAE and \textit{GroupEnc} ($\gamma=4$) with colour labels according to labelled developmental stages of embedded cells.}
    \label{fig_2d}
  \end{framed}
  \vspace{-0.8cm}
\end{figure}

For the Farrell dataset, we also plot the 2-dimensional embeddings from both models and label individual embedded points using annotation provided by the authors (Figure \ref{fig_2d}).
The labels are ordered and correspond to developmental stages of cells in zebrafish embryogenesis.
This is to show the developmental gradient is more apparent in the \textit{GroupEnc} embedding.

\nopagebreak
The results show that, intuitively, both local and global structures in terms of neighbour ranks are preserved worse with decreased dimensionality of the embedding, and this holds across all tested datasets and models (VAE and \textit{GroupEnc} with group sizes of 4, 5 and 6).

Furthermore, the VAE model generally outperforms the \textit{GroupEnc} models when it comes to Local SP.
However, we see consistently better Global SP for \textit{GroupEnc}, concordant with the scale-agnostic nature of the group loss that \textit{GroupEnc} optimises.
Differences between \textit{GroupEnc} models with different group sizes are not significant.

\section{Discussion}

Faithful reconstructions of global relationships in lower-dimensional embeddings are of interest for purposes of visualisation, as well as the potential for downstream processing of data.
We set out to design a deep learning model that uses a loss function for scale-agnostic preservation of randomly sampled structures \cite{Lambert2021}.
We have done this to demonstrate the improvement in global structure preservation (versus VAE) via this loss function and that it can be used in a deep learning context, which has the advantage of providing a parametric model to be trained on a subset of data and used to transform new samples.

The use of geometric priors (similarity matrices, topological priors) with VAEs for dimensionality reduction \cite{Vandaele2022,Kopf2021} is another promising avenue of research in analyses of high-dimensional datasets.
With data that is high-dimensional and noisy by its nature (of which biological single-cell data is an instance), feature engineering by the means of constructing such lower-dimensional embeddings can help extract more salient information about the differential expression of genes in cells, continuous developmental gradients or batch effects between cohorts of samples.

In general, preserving global structures, as opposed to constraining the optimisation process to local structure preservation (as in \textit{t}-SNE \cite{VanDerMaaten2008} or \textit{UMAP} \cite{McInnes2020}) can prove beneficial for analysing hierarchical relationships, developmental gradients and pathways.

Our future work in dimensionality reduction of biological data will focus on effective reconstruction of trajectories, tackling noise and an extended range of evaluation metrics, both unsupervised and supervised.

\section{Code availability}

We make a TensorFlow implementation of \textit{GroupEnc}, including Bash scripts for generating benchmarking jobs (on Slurm) with custom datasets, available at \texttt{github.com/saeyslab/GroupEnc}.

\subsection{Data availability}

We downloaded the Shekhar and Liu datasets via the \textit{scRNAseq} R package \cite{Risso2021} using the functions \texttt{ShekharRetinaData} and \texttt{LiuBrainData} and converted them to \texttt{AnnData} objects using the \textit{scDIOR} \cite{Feng2022} packages for R/Python interoperability.
Other datasets come from the \textit{Single Cell Portal}\footnote[1]{\texttt{https://singlecell.broadinstitute.org/single\_cell}} and are accessible using the following accession numbers.
\begin{itemize}
  \item Farrell: \texttt{SCP162}
  \item Ximerakis: \texttt{SCP263}
  \item Ziegler: \texttt{SCP1289}
  \item Liu: \texttt{SCP2161}
\end{itemize}

\subsection{Data pre-processing}\label{preprocessing}

We used the \textit{scanpy} package version 1.9.1 \cite{Wolf2018} for data pre-processing.
We applied the following Python code for scaling, normalisation and principal component analysis (PCA) prior to running the DR algorithms:

\small{
\begin{verbatim}

import scanpy as sc
hd = sc.pp.normalize_per_cell(X, copy=True) # assume X is count matrix
                                            # (numpy.ndarray)
hd = sc.pp.log1p(hd, copy=True)
hd = sc.pp.scale(hd, max_value=10, copy=True)
data = sc.tl.pca(hd, svd_solver=`arpack', n_comps=50, copy=True)

\end{verbatim}
}

The Farrell dataset was an exception, where already scaled data was used, and only the PCA step remained.

\bibliographystyle{abbrv}
%\bibliography{bibliography}

\begin{thebibliography}{10}

\bibitem{Amid2022}
E.~Amid and M.~K. Warmuth.
\newblock {TriMap}: {Large}-scale {Dimensionality} {Reduction} {Using}
  {Triplets}, Mar. 2022.
\newblock arXiv:1910.00204 [cs, stat].

\bibitem{Amodio2019}
M.~Amodio, D.~van Dijk, K.~Srinivasan, W.~S. Chen, H.~Mohsen, K.~R. Moon,
  A.~Campbell, Y.~Zhao, X.~Wang, M.~Venkataswamy, A.~Desai, V.~Ravi, P.~Kumar,
  R.~Montgomery, G.~Wolf, and S.~Krishnaswamy.
\newblock Exploring single-cell data with deep multitasking neural networks.
\newblock {\em Nature Methods}, 16(11):1139--1145, Nov. 2019.

\bibitem{Chen2020}
L.~Chen, W.~Wang, Y.~Zhai, and M.~Deng.
\newblock Deep soft {K}-means clustering with self-training for single-cell
  {RNA} sequence data.
\newblock {\em NAR Genomics and Bioinformatics}, 2(2):lqaa039, June 2020.

\bibitem{Ding2018}
J.~Ding, A.~Condon, and S.~P. Shah.
\newblock Interpretable dimensionality reduction of single cell transcriptome
  data with deep generative models.
\newblock {\em Nature Communications}, 9(1):2002, May 2018.

\bibitem{Farrell2019}
J.~A. Farrell, Y.~Wang, S.~J. Riesenfeld, K.~Shekhar, A.~Regev, and A.~F.
  Schier.
\newblock Single-cell reconstruction of developmental trajectories during
  zebrafish embryogenesis.
\newblock {\em Science}, 360(6392):eaar3131, June 2018.

\bibitem{Feng2022}
H.~Feng, L.~Lin, and J.~Chen.
\newblock {scDIOR}: single cell {RNA}-seq data {IO} software.
\newblock {\em BMC Bioinformatics}, 23(1):16, Dec. 2022.

\bibitem{Kopf2021}
A.~Kopf, V.~Fortuin, V.~R. Somnath, and M.~Claassen.
\newblock Mixture-of-{Experts} {Variational} {Autoencoder} for clustering and
  generating from similarity-based representations on single cell data.
\newblock {\em PLOS Computational Biology}, 17(6):e1009086, June 2021.

\bibitem{Lambert2021}
P.~Lambert, C.~de~Bodt, M.~Verleysen, and J.~A. Lee.
\newblock {Stochastic quartet approach for fast multidimensional scaling}.
\newblock pages 417--422, 2021.

\bibitem{Lee2015}
J.~A. Lee, D.~H. Peluffo-Ordóñez, and M.~Verleysen.
\newblock Multi-scale similarities in stochastic neighbour embedding:
  {Reducing} dimensionality while preserving both local and global structure.
\newblock {\em Neurocomputing}, 169:246--261, Dec. 2015.

\bibitem{Liu2023}
Y.~Liu, E.~L. Savier, V.~J. DePiero, C.~Chen, D.~C. Schwalbe, R.-J.
  Abraham-Fan, H.~Chen, J.~N. Campbell, and J.~Cang.
\newblock Mapping visual functions onto molecular cell types in the mouse
  superior colliculus.
\newblock {\em Neuron}, 111(12):1876--1886.e5, 2023.

\bibitem{McInnes2020}
L.~McInnes, J.~Healy, and J.~Melville.
\newblock {UMAP}: {Uniform} {Manifold} {Approximation} and {Projection} for
  {Dimension} {Reduction}, Sept. 2020.
\newblock arXiv:1802.03426 [cs, stat].

\bibitem{Risso2021}
D.~Risso and M.~Cole.
\newblock {\em {scRNAseq}: {Collection} of {Public} {Single-Cell} {RNA-Seq}
  {Datasets}}, 2021.
\newblock R package version 2.8.0.

\bibitem{Shekhar2016}
K.~Shekhar, S.~W. Lapan, I.~E. Whitney, N.~M. Tran, E.~Z. Macosko,
  M.~Kowalczyk, X.~Adiconis, J.~Z. Levin, J.~Nemesh, M.~Goldman, S.~A.
  McCarroll, C.~L. Cepko, A.~Regev, and J.~R. Sanes.
\newblock Comprehensive {Classification} of {Retinal} {Bipolar} {Neurons} by
  {Single}-{Cell} {Transcriptomics}.
\newblock {\em Cell}, 166(5):1308--1323.e30, Aug. 2016.

\bibitem{Szubert2019}
B.~Szubert, J.~E. Cole, C.~Monaco, and I.~Drozdov.
\newblock Structure-preserving visualisation of high dimensional single-cell
  datasets.
\newblock {\em Scientific Reports}, 9(1):8914, June 2019.

\bibitem{VanDerMaaten2008}
L.~van~der Maaten and G.~Hinton.
\newblock Visualizing {Data} using {t-SNE}.
\newblock {\em Journal of Machine Learning Research}, 9(86):2579--2605, 2008.

\bibitem{Vandaele2022}
R.~Vandaele, B.~Kang, J.~Lijffijt, T.~De~Bie, and Y.~Saeys.
\newblock Topologically {Regularized} {Data} {Embeddings}, Mar. 2022.
\newblock arXiv:2110.09193 [cs, stat].

\bibitem{Wolf2018}
F.~A. Wolf, P.~Angerer, and F.~J. Theis.
\newblock {SCANPY}: large-scale single-cell gene expression data analysis.
\newblock {\em Genome Biology}, 19(1):15, Dec. 2018.

\bibitem{Ximerakis2019}
M.~Ximerakis, S.~L. Lipnick, B.~T. Innes, S.~K. Simmons, X.~Adiconis,
  D.~Dionne, B.~A. Mayweather, L.~Nguyen, Z.~Niziolek, C.~Ozek, V.~L. Butty,
  R.~Isserlin, S.~M. Buchanan, S.~S. Levine, A.~Regev, G.~D. Bader, J.~Z.
  Levin, and L.~L. Rubin.
\newblock Single-cell transcriptomic profiling of the aging mouse brain.
\newblock {\em Nature Neuroscience}, 22(10):1696--1708, Oct. 2019.

\bibitem{Ziegler2021}
C.~G. Ziegler, V.~N. Miao, A.~H. Owings, A.~W. Navia, Y.~Tang, J.~D. Bromley,
  P.~Lotfy, M.~Sloan, H.~Laird, H.~B. Williams, M.~George, R.~S. Drake,
  T.~Christian, A.~Parker, C.~B. Sindel, M.~W. Burger, Y.~Pride, M.~Hasan,
  G.~E. Abraham, M.~Senitko, T.~O. Robinson, A.~K. Shalek, S.~C. Glover, B.~H.
  Horwitz, and J.~Ordovas-Montanes.
\newblock Impaired local intrinsic immunity to {SARS-CoV-2} infection in severe
  {COVID-19}.
\newblock {\em Cell}, 184(18):4713--4733.e22, 2021.

\end{thebibliography}

\end{document}